\pdfoutput=1

\documentclass[11pt]{article}

\usepackage[]{acl}

\usepackage{times}
\usepackage{latexsym}

\usepackage[T1]{fontenc}

\usepackage[utf8]{inputenc}

\usepackage{microtype}
\usepackage{graphicx}
\usepackage{amsmath} 
\usepackage{amsfonts}
\usepackage{booktabs}
\usepackage{subfig}

\title{Cross-domain Named Entity Recognition via Graph Matching}

\author{Junhao Zheng, Haibin Chen, Qianli Ma*\\
  School of Computer Science and Engineering, \\
  South China University of Technology, Guangzhou, China\\
  \texttt{junhaozheng47@outlook.com}, \texttt{haibin\_chen@foxmail.com}, \\
  \texttt{qianlima@scut.edu.cn}\thanks{*Corresponding author}}
\begin{document}
\maketitle
\begin{abstract}
Cross-domain NER is a practical yet challenging problem since the data scarcity in the real-world scenario. 
A common practice is first to learn a NER model in a rich-resource general domain and then adapt the model to specific domains.
Due to the mismatch problem between entity types across domains, the wide knowledge in the general domain can not effectively transfer to the target domain NER model.
To this end, we model the label relationship as a probability distribution and construct label graphs in both source and target label spaces.
To enhance the contextual representation with label structures, we fuse the label graph into the word embedding output by BERT.
By representing label relationships as graphs, we formulate cross-domain NER as a graph matching problem.
Furthermore, the proposed method has good applicability with pre-training methods and is potentially capable of other cross-domain prediction tasks.
Empirical results on four datasets show that our method outperforms a series of transfer learning, multi-task learning, and few-shot learning methods.
\end{abstract}

\section{Introduction}
Named entity recognition (NER) is a crucial component in many language understanding tasks \citep{DBLP:journals/coling/Shaalan14,nadeau2007survey} and is often applied in various domains. 
Due to the data scarcity in the real-world scenario, obtaining adequate domain-specific data is usually expensive and time-consuming. 
Hence, cross-domain NER, which is capable of adapting NER models to specific domains with limited data, has been drawing increasing attention in recent years.

However, one of the primary challenges of cross-domain NER is the mismatch between source and target domain labels \citep{DBLP:conf/emnlp/YangK20}. 
For example, the label sets between ATIS  \citep{DBLP:conf/interspeech/Hakkani-TurTCCG16} and CoNLL 2003 are non-overlapping.
To address this issue, some approaches utilize multi-task learning  \citep{DBLP:conf/acl/JiaZ20,DBLP:conf/acl/WangKP20} for transferring knowledge across domains.
However, these methods require full training on both source and target domain data when adapting to each new domain.
Since the source domain dataset is usually much larger than the target domain dataset, the multi-task learning methods are inefficient when adapting to low-resource domains.

\begin{figure}[t]
    \centering
    \includegraphics[width=\linewidth]{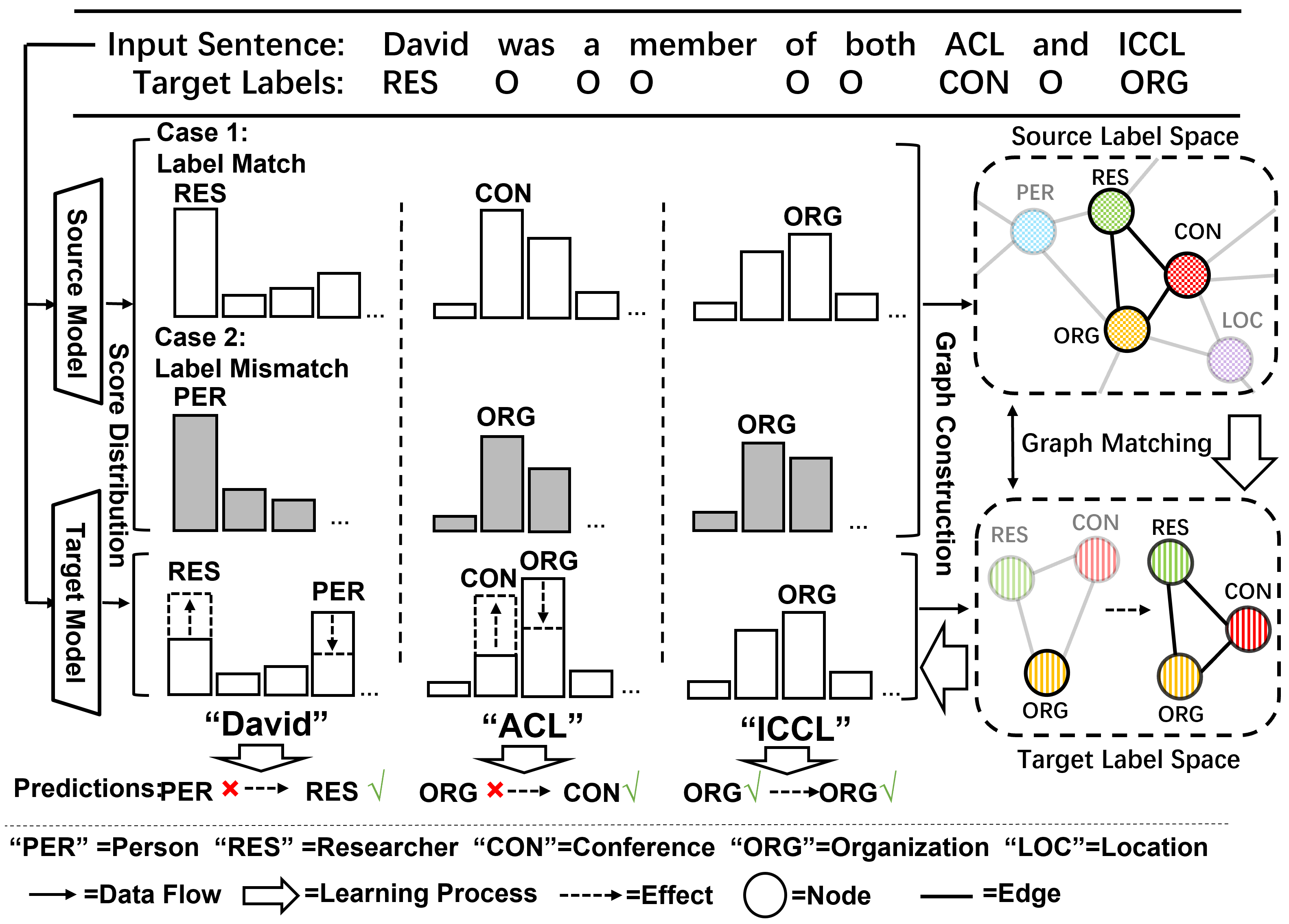}
    \caption{A demonstration of graph matching. In both two cases, our model learns graph structures from the source label space and makes correct predictions. In two label spaces, each node is a target label and the matching nodes and edges are opaque.}
    \label{fig:fig1}
\end{figure}

Recently, as Pre-trained Language Models (PLMs) such as BERT \citep{DBLP:conf/naacl/DevlinCLT19} have shown remarkable success in NER, transfer-learning-based methods also show effectiveness for cross-domain NER.
A typical approach is to first train a NER model initialized with PLM on rich-resource domain  (\textit{e.g.,} CoNLL 2003 \citep{DBLP:conf/conll/SangM03}), and then fine-tune the entire model with a new task-specific linear classifier (\textit{pre-train fine-tune}) \citep{DBLP:conf/lrec/LeeDS18,DBLP:conf/coling/RodriguezCL18}.
Despite its simplicity, this approach provides strong results on several benchmarks \citep{huang2020few}, and we serve it as the baseline in our research.

Inspired by the idea in \citet{DBLP:conf/nips/YouKL020}, where labels across domains are connected by probability distributions, we propose a novel approach, \textbf{L}abel \textbf{S}tructure \textbf{T}ransfer for cross-domain \textbf{NER} (\textbf{LST-NER}) to address the label mismatch problem.
By modeling the label relationships as \textit{label graphs}, we transfer the label structure from the source model (\textit{i.e.,} the NER model trained on source domain) to the target model (\textit{i.e.,} fine-tuned model).
We are the first to capture label graph structures for cross-domain NER to our best knowledge.
In this study, we focus on enhancing cross-domain ability based on \textit{pretrain-finetune} training paradigm, with \textbf{only} target domain labeled data for domain adaptation.
Therefore, pre-training \citep{DBLP:conf/aaai/Liu0YDJCMF21} and self-training \citep{huang2020few} based methods, which leverages massive unlabeled data, are not considered.

To explicitly capture the connections between two domains labels, we construct a \textit{label graph} by probability distributions of target labels estimated by the source NER model.
In the \textit{label graph}, graph nodes refer to target labels, and edges refer to the relationships between labels.
We represent each node as the probability distribution and add an edge between two nodes if the labels have similar distributions.
By representing label relationships as \textit{label graphs} in both source and target label spaces, the label knowledge can be transferred via graph matching.
We introduce Gromov-Wasserstein distance (GWD) for aligning two \textit{label graphs} because of its capability of capturing edge similarity.

We show an example in Fig \ref{fig:fig1} to demonstrate how graph matching works.
In the example, "ACL" is a "Conference" named entity in the target domain. 
When label sets between source and target domains match perfectly, the source NER model naturally predicts "Conference" with the highest probability.
Then, the target model straightforward learns this property from the source domain.
When two label sets are mismatching, the source NER model may predict "ACL" as an "Organization" since the label "Organization" is seen in the source domain.
By score distributions of "ICCL" and "David" in the source domain, we can model their relationships with "ACL" as graph structures.
Then, the target model learns label structures via graph matching and predicts "ACL" as "Conference" correctly.
In this way, the label relationships can be learned even when two domain label sets are different.

Furthermore, we enhance the contextual representation by fusing the constructed \textit{label graph} into the word embedding by Graph Convolutional Network (GCN), where an auxiliary task is introduced for better extracting label-specific components for each entity type.

We performed extensive experiments on eight different domains in both rich- and low-resource settings.
Empirical results show that our method outperforms a series of competitive baselines.

\section{Related Work}
\textbf{Cross-domain NER.} \quad 
In recent years, cross-domain NER has received increasing research attention. 
There is a line of research based on multi-task learning \citep{DBLP:conf/iclr/YangSC17}.
Some approaches proposed adding auxiliary tasks \citet{DBLP:conf/rep4nlp/LiuWF20,DBLP:conf/acl/WangKP20}, while some approaches proposed new model architecture \citep{DBLP:conf/acl/JiaXZ19,DBLP:conf/acl/JiaZ20} for improving target domain NER model by jointly training on both source and target domain data.
\citet{DBLP:conf/acl/JiaZ20} presented a multi-cell compositional LSTM (Multi-Cell LSTM) structure where modeled each entity type as a separate cell state, and it reaches the state-of-the-art (SOTA) performance for cross-domain NER.
These methods require training on massive source domain data when adapting to each domain and thus inefficient.

Another line of research is based on transfer learning.
Except from the \textit{pretrain-finetune} paradigm, some approaches proposed adding adaption layers \citet{DBLP:conf/emnlp/LinL18} or adapter modules \citet{DBLP:conf/icml/HoulsbyGJMLGAG19} to the backbone network.
Compared with them, our method constructs \textit{label graphs} dynamically and performs label semantic fusion via attention mechanism, and thus has fewer parameters for training.
Besides, our method is built on word contextual embedding by PLM.
Therefore, our model can combine with various backbone networks and thus has better applicability.

\noindent\textbf{Few-shot NER.} \quad 
Few-shot NER aims at recognizing new categories in a highly low-resource scenario \citep{DBLP:conf/ijcai/FengF0F018}, which also shows good cross-domain ability.
\citet{tong-etal-2021-learning} induced different undefined classes from the "Others" class \cite{zheng2022distilling} to alleviate the over-fitting problem.
\citet{DBLP:conf/emnlp/YangK20} proposed NNShot and StructShot based on the nearest neighbor classifier, and StructShot further applies the Viterbi algorithm when decoding.
The few-shot learning methods focus on building models that can generalize from very few examples. 
Unlike these methods, our approach aims to enhance domain adaptation ability in both low-resource and rich-resource scenarios.

\section{Methodology}

\subsection{Problem Formulation}\label{problem_formulation}
We focus on only one source and one target domain in this study. 
Given a NER model $f_0$ pre-trained on a source dataset $\mathcal{D}_s=\{(x_s^i,y_s^i)\}_{i=1}^{m_s}$, we aims to fine-tune $f_0$ by a target dataset $\mathcal{D}_t=\{(x_t^i,y_t^i)\}_{i=1}^{m_t}$. 
Following \citet{DBLP:conf/nips/YouKL020}, we assume that only $\mathcal{D}_t$ and $f_0$ are available when fine-tuning since $\mathcal{D}_s$ is often large-scale. 

\begin{figure*}[!t]
    \centering
    \includegraphics[width=\linewidth]{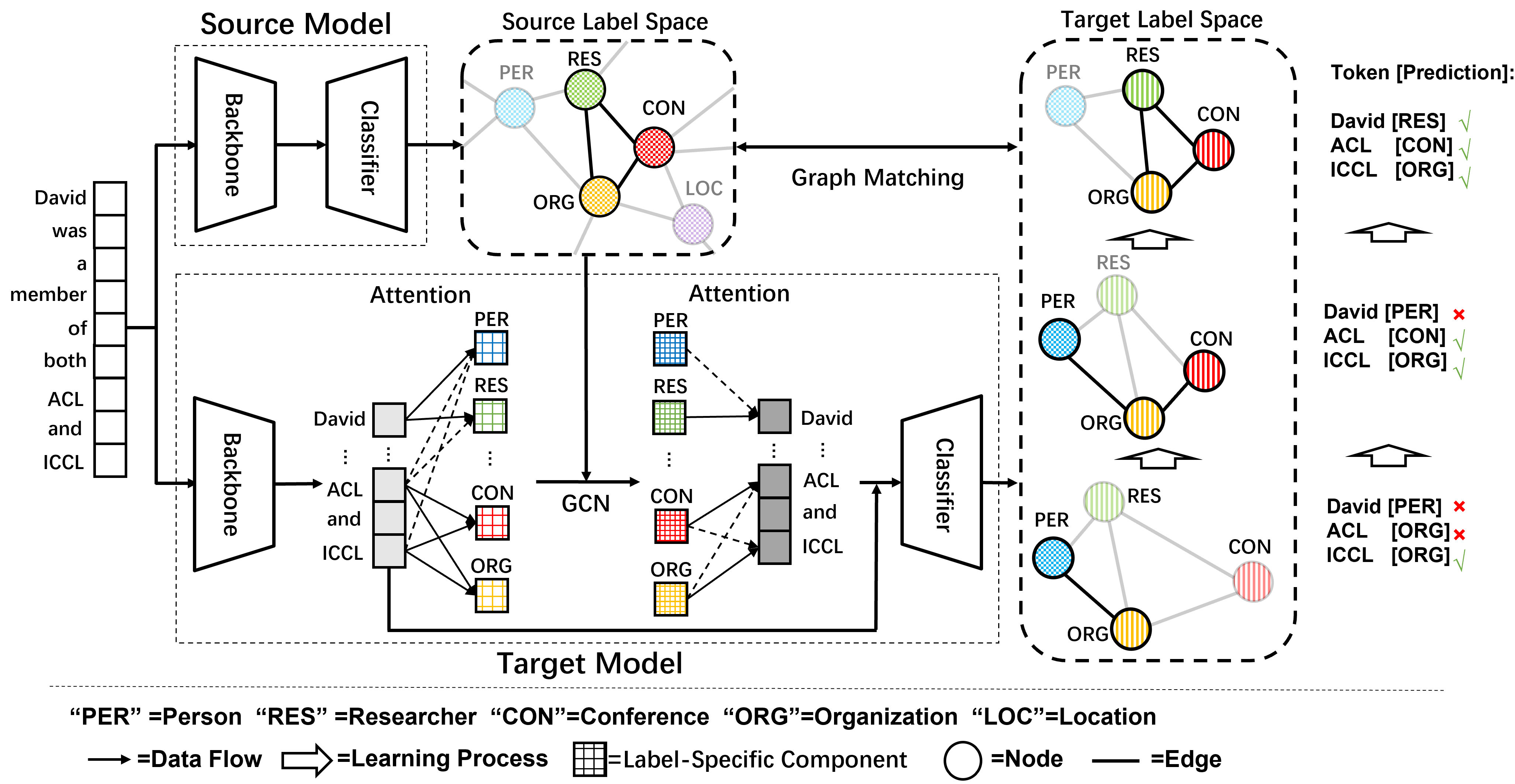}
    \caption{A demonstration of the proposed model. First, the label graph from source label space is incorporated into the contextual representation by GCN. Then, the target model transfers graph structures from the source model via graph matching. Finally, the target model makes correct predictions with the learned label structures.}
    \label{fig:fig2}
\end{figure*}

Because the source label set $\mathcal{Y}_s$ and target label set $\mathcal{Y}_t$ may be mismatching, $f_0$ can not be applied to target data directly. 
A common practice is to split $f_0$ into two parts: a backbone network for learning general representation and task-specific layers for mapping representation to label space. 
We adopt BERT as our backbone model and Fully-Connected (FC) layer as the task-specific layer throughout our research.
We show a demonstration of our model in Fig. \ref{fig:fig2}.

\subsection{Label Graph Construction}\label{build_graph}
In this part, we construct \textit{source graph} and \textit{target graph} with the probabilistic outputs of source and target NER models respectively.

Typically, the target labels are fine-grained and domain-specific, while the source labels are coarse-grained and more general. 
Similar to the idea of \citet{DBLP:conf/nips/YouKL020}, we map each target label as a probability distribution of the source labels.
A straightforward method for obtaining this mapping (\textit{i.e.} conditional distribution) $p(y_s|y_t=y)$ is to average the predictions of the source model over all samples for each target entity type. Formally, we have
\begin{equation}
\begin{aligned}
    p(y_s|y_t=y) \approx |\mathcal{D}^y_t|^{-1}\Sigma_{(x_t,y_t)\in \mathcal{D}^y_t} f_0'(x_t) \\
    f_0'(x_t) = softmax(f_0(x_t)/T) \\
    \mathcal{D}^y_t = \{(x_t,y_t)\in \mathcal{D}_t | y_t=y \}, \\
\end{aligned}
\end{equation}
where $y_s$/$y_t$ denotes source/target label, $T$ denotes the temperature parameter for smoothing the probability distribution and $|\mathcal{D}^y_t|$ is the number of target domain training samples \footnote{The training sample refers to one token and its ground-truth label} with ground-truth label $y$.
The pre-trained model $f_0$ is regarded as a probabilistic model for approximating the probability distribution $p(y_s|x_t)$ over source labels $\mathcal{Y}_s$.

Next, we build \textit{source graph} $\mathcal{G}_s(\mathcal{V}_s,\mathcal{E}_s)$ where nodes refer to target labels and edges refer to semantic similarity between nodes.
As illustrated in Fig \ref{fig:fig1}, two labels with similar semantic meanings (\textit{i.e.,} "Conference" and "Organization") have the similar probability distribution.
Based on this feature, we represent the graph node with label $y$ as
\begin{equation}
    \begin{aligned}
    \widetilde{\boldsymbol{v}}_{s}^y = \left[ p(y_s^{(1)}|y_t=y),\cdots,p(y_s^{(i)}|y_t=y) \right] \\
     y_s^{(i)} \in \mathcal{Y}_s, i \in \{1,\cdots,|\mathcal{Y}_s|\}
    \end{aligned}
\end{equation}
where $\widetilde{\boldsymbol{v}}_{s}^y \in \mathbb{R}^{|\mathcal{Y}_s|}$ is the node representation and $|\mathcal{Y}_s|$ is the number of source labels.
To eliminate the influence of scales of different dimensions, we normalize the graph nodes by dividing the average distance of node pairs, and $l2$ distance is used as the distance metric.
Then, the graph node representation for label $y$ is calculated as
\begin{equation}
    \boldsymbol{v}_s^y = \frac{\widetilde{\boldsymbol{v}}_{s}^y*|\mathcal{Y}_t|^2}{\sum_{y_1,y_2}l2(\widetilde{\boldsymbol{v}}_{s}^{y_1},\widetilde{\boldsymbol{v}}_{s}^{y_2})},
\end{equation}
where $\boldsymbol{v}_s^y\in \mathcal{R}^{|\mathcal{Y}_s|}$ is the normalized node representation, $|\mathcal{Y}_t|$ is the number of target labels and $l2$ is the distance function.
Then, we add edge between two nodes if and only if their distance is smaller than a threshold $\delta$.
\begin{equation}
    \boldsymbol{e}_s^{y_1,y_2} = 
    \begin{cases}
    l2(\boldsymbol{v}_s^{y_1},\boldsymbol{v}_s^{y_2}),  &\text{if } l2(\boldsymbol{v}_s^{y_1},\boldsymbol{v}_s^{y_2})<\delta;\\
    \infty, &\text{else}.\\
    \end{cases}
\end{equation}

In a similar way as \textit{source graph}, we construct \textit{target graph} $\mathcal{G}_t(\mathcal{V}_t,\mathcal{E}_t)$ by the fine-tuned model $f$ where probability distribution $p(y_t|x)$ over target labels $\mathcal{Y}_t$ are estimated.
In \textit{target graph}, nodes refer to target labels, and edges refer to semantic similarity measured in target label space.

\subsection{Label Semantics Fusion}\label{semantic_learning}
Commonly in NER, the ground-truth label of a named entity is related to the context (\textit{e.g.,} label "Researcher" can be inferred by label "Conference" as the example shown in Fig \ref{fig:fig1}). 
In this part, we fused the learned graph structure into the word contextual embedding output by BERT to model the sentence's semantic label relationships.

Given a sentence $X = [x_1,\cdots, x_{n_s}]$ with ground-truth label sequence $Y$, the contextual representation $\boldsymbol{h}_j\in\mathbb{R}^{d_h}$ for each token can be obtained by backbone network.
Then, we randomly initialize the \textit{label representation} $\boldsymbol{c}_i \in \mathbb{R}^{d_c}$ before fine-tuning.
The \textit{label representation} represents the semantic meanings for each entity type, and it is learned during fine-tuning.
For the sentence $X$, we apply a label-guided attention mechanism to extract the \textit{label-specific components} as follow:
\begin{equation}
\begin{aligned}
    \boldsymbol{q}_j &= \boldsymbol{h}_j\mathbf{W}_p+\boldsymbol{b}_p, \\
    \alpha_{ij} &= \frac{\exp(\boldsymbol{q}_j \boldsymbol{c}_i^T)}
                {\sum_j{\exp(\boldsymbol{q}_j \boldsymbol{c}_i^T)} }, \\
    \boldsymbol{u}_i &= \sum_j{\alpha_{ij}\boldsymbol{q}_j}, \\
\end{aligned}
\end{equation}
where $\boldsymbol{q}_j \in \mathbb{R}^{d_p}$ is the \textit{label-related embedding} for the $j$-th token in the sentence, $\mathbf{W}_p \in \mathbb{R}^{d_h \times d_p}$ and $\boldsymbol{b}_p \in \mathbb{R}^{d_p}$ are the weight and bias for projection respectively.
$\boldsymbol{u}_i\in\mathbb{R}^{d_p}$ denotes the \textit{label-specific component} for the $i$-th label in $\mathcal{Y}_t$ and $\alpha_{ij}$ indicates how informative the $j$-th token to the $i$-th label.
For each sentence, \textit{label-specific components} modeling its semantic relevance to each entity type.

And then, by replacing the node representation of \textit{source graph} from probability distribution $\boldsymbol{v}_s$  to \textit{label-specific component} $\boldsymbol{u}$, we obtain the graph representation of \textit{label-specific components}.
Next, we utilize GCN \citep{DBLP:conf/iclr/KipfW17} to enhance the representations of each \textit{label-specific component} by propagating messages between neighboring nodes.
\begin{equation}
    \boldsymbol{u}' = \textit{GCN}(\boldsymbol{u})
\end{equation}
$\boldsymbol{u}'\in\mathbb{R}^{d_p}$ denotes the aggregated node representation of the \textit{label-specific component} and \textit{GCN} denotes the graph convolution operations where details are omitted for simplicity.
As shown in Fig. \ref{fig:fig2}, label structure from \textit{source graph} is fused into \textit{label-specific components} by GCN.

Last, we utilize the token-guided attention mechanism to fuse the aggregated \textit{label-specific component} into the contextual representation for each word:
\begin{equation}\label{eq:label_fusion}
\begin{aligned}
    \beta_{ji} &= \frac{\exp(\boldsymbol{q}_j \boldsymbol{u}_i^{'T})}
                {\sum_i{\exp(\boldsymbol{q}_j \boldsymbol{u}_i^{'T})} } \\
    \boldsymbol{h}'_j &= \boldsymbol{h}_j + (\sum_i{\beta_{ji}\boldsymbol{u}'_i})\mathbf{W}'_p+\mathbf{b}'_p. \\
\end{aligned}
\end{equation}
$\boldsymbol{h}'_j\in\mathbb{R}^{d_h}$ is the \textit{label-fused embedding} for the $j$-th token and $\mathbf{W}'_p\in \mathbb{R}^{d_p\times d_h}$,$\mathbf{b}'_p\in \mathbb{R}^{d_h}$ are the weight and bias for projection respectively.
In Eq. \ref{eq:label_fusion}, we map the weighted sum of $\boldsymbol{u}'$ into the same space of $\boldsymbol{h}_j$ and add them together to allow information fusion. 
Followed by the task-specific FC layer, the classification loss for NER tasks can be calculated:
\begin{equation}
    \mathcal{L}_{cls} = \textit{CE}(\textit{FC}(\boldsymbol{h}'),Y)
\end{equation}
where \textit{CE} denotes the Cross-Entropy loss.

Besides, we introduce an auxiliary task to ensure the \textit{label-specific components} focus on correct entity types.
Concretely, the model predicts what entity types appear in the sentence, which is a multi-label classification task.
The loss for the auxiliary task is calculated as
\begin{equation}
    \mathcal{L}_{aux} = \textit{BCE}(\textit{FC}_{aux}(\textit{Cat}([\boldsymbol{h}'_1,\cdots,\boldsymbol{h}'_{n_s}])),Y')
\end{equation}
where \textit{BCE} is the Binary-Cross-Entropy loss, $\textit{FC}_{aux}$ is the FC layer for auxiliary task, $\textit{Cat}$ is the concatenation operation for last dimension and $Y'$ is the ground-truth label for the sentence. 
Different from $\mathcal{L}_{cls}$, $\mathcal{L}_{aux}$ encourages model to extract correct \textit{label-specific components} for each sentence. 

\subsection{Graph Structure Matching}\label{graph_matching}
Since \textit{source graph} $\mathcal{G}_s$ is constructed by the pre-trained LM $f_0$, it naturally contain priori knowledge from rich-resource domain.
In this part, we utilize the \textit{label graphs} built in different label spaces for graph matching to exploit the semantic relations among labels from \textit{source graph}.

Gromov-Wasserstein distance (GWD) is proposed for distributional metric matching by \citet{DBLP:conf/icml/PeyreCS16}.
Since its capability of capturing edge similarity between graphs, GWD has been applied to graph matching \citep{DBLP:conf/icml/VayerCTCF19,chowdhury2019gromov} and domain alignment \citep{DBLP:conf/icml/ChenG0LC020}.
Naturally, we can adopt GWD for matching the edges (relationships) between two \textit{label graphs}.

Following \citet{DBLP:conf/emnlp/Alvarez-MelisJ18,DBLP:conf/icml/ChenG0LC020}, we convert each graph to a discrete distribution with uniform mass on each node.
Let $\boldsymbol{\mu}$,$\boldsymbol{\nu}$ denote two discrete distributions corresponding to $\mathcal{G}_s,\mathcal{G}_t$ respectively. 
Then, we define the GWD between $\boldsymbol{\mu}$ and $\boldsymbol{\nu}$ as: 
\begin{equation}\label{eq:objective_gwd}
\begin{aligned}
& \mathbf{D}_{gw}(\boldsymbol{\mu}, \boldsymbol{\nu}) \\
&=\inf_{\boldsymbol{\gamma}\in\prod(\boldsymbol{\mu}, \boldsymbol{\nu})} 
\mathbb{E}_{(\boldsymbol{v}_s,\boldsymbol{v}_t)\sim \boldsymbol{\gamma}, (\boldsymbol{v}_s',\boldsymbol{v}_t')\sim \boldsymbol{\gamma}}[L(\boldsymbol{v}_s,\boldsymbol{v}_t,\boldsymbol{v}_s',\boldsymbol{v}_t')] \\
&=\min_{\hat{\mathbf{T}}\in\prod(\boldsymbol{u}, \boldsymbol{v})}
\sum_{i,i',j,j'}\hat{\mathbf{T}}_{ij}\hat{\mathbf{T}}_{i'j'}
L(\boldsymbol{v}_s^i,\boldsymbol{v}_t^j,\boldsymbol{v}_s^{i'},\boldsymbol{v}_t^{j'}),
\end{aligned}
\end{equation}
where $\prod(\boldsymbol{\mu},\boldsymbol{\nu})$ denotes all the joint distributions $\boldsymbol{\gamma}(\boldsymbol{v}_s,\boldsymbol{v}_t)$ with marginals $\boldsymbol{\mu}(\boldsymbol{v}_s)$ and $\boldsymbol{\nu}(\boldsymbol{v}_t)$.
$\prod(\boldsymbol{u},\boldsymbol{v})$ represents the space of all valid transport plan, where the weight vector $\boldsymbol{u}=\{u_i\}_{i=1}^n,\boldsymbol{v}=\{v_i\}_{i=1}^m$ is the $n$- and $m$-dimensional simplex for distribution $\boldsymbol{\mu}$,$\boldsymbol{\nu}$.
The matrix $\mathbf{T}$ is the transport plan, where $\mathbf{T}_{ij}$ represents the amount of mass shifted from $u_i$ to $v_j$.
$L(\cdot)$ is the cost function evaluating the intra-graph structural similarity between two pairs of nodes $(\boldsymbol{v}_s^i,\boldsymbol{v}_s^{i'})$ and $(\boldsymbol{v}_t^j,\boldsymbol{v}_t^{j'})$, and it is defined as follow in the proposed method:
\begin{equation}
    L(\boldsymbol{v}_s^i,\boldsymbol{v}_t^j,\boldsymbol{v}_s^{i'},\boldsymbol{v}_t^{j'})=|l2(\boldsymbol{v}_s^i,\boldsymbol{v}_s^{i'}) - l2(\boldsymbol{v}_t^j,\boldsymbol{v}_t^{j'}) |
\end{equation}

By projecting the edges into nodes, the learned transport plan $\hat{\mathbf{T}}$ helps align the edges in different graphs \citep{DBLP:books/daglib/0073486}.
Then, label relationships (edges) can be learned from \textit{source graph} to \textit{target graph} by minimizing $\mathbf{D}_{gw}$ with Sinkhorn algorithm \citep{DBLP:conf/nips/Cuturi13,peyre2019computational}.
In Fig. \ref{fig:fig2}, the fine-tuned model learns the structure between labels (\textit{i.e.,} "Conference", "Organization" and "Researcher") , and makes correct predictions with the learned label relationships. 
When fine-tuning, \textit{target graph} evolves dynamically through the update of the parameters of NER model $f$, while \textit{source graph} and the source model $f_0$ are frozen.

\subsection{Total Learning Objective}\label{learning_objective}
Finally, the total loss can be formulated as
\begin{equation}
    \mathcal{L} = \mathcal{L}_{cls} + \lambda_1 \mathcal{L}_{aux} + \lambda_2 \mathbf{D}_{gw},
\end{equation}
where the loss of auxiliary task and GWD are weighted by $\lambda_1$ and $\lambda_2$ respectively. 

\begin{table*}[!t]
\tiny
\centering
\resizebox{\linewidth}{!}{
    \begin{tabular}{c|c|c|c|c|ccccc}
    \toprule
    \textbf{Datasets} & CoNLL 2003 & MIT Movie & MIT Restaurant & ATIS  & \multicolumn{5}{c}{CrossNER} \\
    \midrule
    \textbf{Domain} & News  & Movie Reviews & Restaurant Reviews & Dialogue & Politics & Natural Science & Music & Literature & Artificial Intelligence \\
    \textbf{\#Train} & 15.0k & 7.8k  & 7.7k  & 5.0k  & 200   & 200   & 100   & 100   & 100 \\
    \textbf{\#Test} & 3.7k  & 2.0k  & 1.5k  & 893   & 651   & 543   & 456   & 416   & 431 \\
    \textbf{\#Entity Type} & 4     & 12    & 8     & 79    & 10    & 17    & 13    & 11    & 12 \\
    \bottomrule
    \end{tabular}%
    }
    \caption{Statistics on the 5 public datasets in our experiments}
    \label{tab:dataset_statistics}
\end{table*}

\begin{table*}[!ht]
\tiny
  \centering
  \resizebox{\linewidth}{!}{
    \begin{tabular}{l|p{0.35cm}p{0.35cm}p{0.35cm}p{0.35cm}p{0.35cm}p{0.35cm}p{0.35cm}p{0.5cm}|p{0.35cm}p{0.35cm}p{0.35cm}p{0.35cm}p{0.35cm}p{0.35cm}p{0.35cm}p{0.35cm}}
    \toprule
    \textbf{Samples} & \multicolumn{8}{c|}{\textbf{K=20}}                            & \multicolumn{8}{c}{\textbf{K=50}} \\
    \midrule
    \textbf{Domain} & \textbf{Pol.} & \textbf{Sci.} & \textbf{Mus.} & \textbf{Lit.} & \textbf{AI} & \textbf{Mov.} & \textbf{Res.} & \textbf{Dia.} & \textbf{Pol.} & \textbf{Sci.} & \textbf{Mus.} & \textbf{Lit.} & \textbf{AI} & \textbf{Mov.} & \textbf{Res.} & \textbf{Dia.} \\
    \midrule
    BiLSTM-CRF & 41.75  & 42.54  & 37.96  & 35.78  & 37.59  & 49.98  & 49.65  & 92.32  & 53.46  & 48.89  & 43.65  & 41.54  & 44.73  & 56.13  & 58.11  & 94.28  \\
    BiLSTM-CRF-joint $\dag$ & 44.62  & 44.91  & 42.28  & 39.54  & 41.23  & 51.73  & 50.61  & 92.54  & 55.17  & 49.68  & 44.58  & 43.14  & 46.35  & 57.60  & 58.94  & 94.58  \\
    Coach $\dag$ & 46.15  & 48.71  & 43.37  & 41.64  & 41.55  & 45.83  & 49.56  & 92.74  & 60.97  & 52.03  & 51.56  & 48.73  & 51.15  & 56.09  & 57.50  & 94.69  \\
    Multi-Cell LSTM $\dag$ & 59.58  & 60.55  & 67.12  & 63.92  & 55.39  & 53.59  & 52.18  & 90.36  & 68.21  & 65.78  & 70.47  & 66.85  & 58.67  & 58.48  & 60.57  & 92.78  \\
    \midrule
    BERT-tagger & 61.01  & 60.34  & 64.73  & 61.79  & 53.78  & 53.39  & 55.13  & 92.48  & 66.13  & 63.93  & 68.41  & 63.44  & 58.93  & 58.16  & 60.58  & 94.51  \\
    BERT-tagger-joint $\dag$ & 61.61  & 60.58  & 64.16  & 60.36  & 53.18  & 53.62  & 55.54  & 91.24  & 66.30  & 64.04  & 67.71  & 62.58  & 58.52  & 58.04  & 60.71  & 93.78  \\
    NNShot & 60.93  & 60.67  & 64.21  & 61.64  & 54.27  & 52.97  & 55.23  & 91.65  & 66.33  & 63.78  & 67.94  & 63.19  & 59.17  & 57.34  & 60.26  & 93.86  \\
    StructShot & 63.31  & 62.95  & 67.27  & 63.48  & 55.16  & 54.83  & 55.93  & 92.66  & 67.16  & 64.52  & 70.21  & 65.33  & 59.73  & 58.74  & 61.60  & 94.38  \\
    \midrule
    templateNER & 63.39  & 62.64  & 62.00  & 61.84  & 56.34  & 40.15  & 47.82  & 58.39  & 65.23  & 62.84  & 64.57  & 64.49  & 56.58  & 43.42  & 54.05  & 59.67  \\
    \midrule
    LST-NER w/o $\mathcal{D}_{gw}$+$\mathcal{L}_{aux}$ & 60.56  & 60.72  & 65.10  & 62.26  & 54.02  & 53.18  & 55.35  & 91.43  & 65.95  & 63.76  & 68.77  & 64.22  & 58.72  & 58.41  & 60.54  & 94.44  \\
    LST-NER w/o $\mathcal{L}_{aux}$ & 62.91  & 62.55  & 66.98  & 63.73  & 56.31  & 56.11  & 57.32  & 92.66  & 68.19  & 64.42  & 70.17  & 66.13  & 59.86  & 60.33  & 62.73  & 94.74  \\
    LST-NER w/o $\mathcal{D}_{gw}$ & 62.16  & 62.39  & 66.28  & 63.85  & 55.82  & 55.27  & 56.92  & 92.87  & 67.63  & 64.94  & 69.76  & 65.24  & 59.12  & 59.56  & 62.21  & 94.59  \\
    LST-NER (Ours) & \textbf{64.06 } & \textbf{64.03 } & \textbf{68.83 } & \textbf{64.94 } & \textbf{57.78 } & \textbf{57.83 } & \textbf{58.26 } & \textbf{93.21 } & \textbf{68.51 } & \textbf{66.48 } & \textbf{72.04 } & \textbf{66.73 } & \textbf{60.69 } & \textbf{61.25 } & \textbf{63.58 } & \textbf{94.94 } \\[-3pt]
    \bottomrule
    \end{tabular}
    }
  \caption{Cross domain results on eight different domains in low-resource setting. $\dag$ indicates both source and target labeled samples are used when training.}
  \label{tab:low_cross_domain_table}%
\end{table*}%

\begin{table}[!ht]
  \tiny
  \centering
  \resizebox{\linewidth}{!}{
    \begin{tabular}{lp{0.7cm}p{0.7cm}p{0.7cm}}
    \toprule
    \textbf{Domain} & \multicolumn{1}{l}{\textbf{Mov.}} & \multicolumn{1}{l}{\textbf{Res.}} & \multicolumn{1}{l}{\textbf{Dia.}} \\
    \midrule
    BiLSTM-CRF & 67.16  & 77.49  & 95.10  \\
    BiLSTM-CRF-joint $\dag$ & 68.31  & 78.13  & 95.26  \\
    Coach $\dag$ & 67.62  & 77.82  & 95.04  \\
    Multi-Cell LSTM $\dag$ & 69.41  & 78.67  & 93.95  \\
    \midrule
    BERT-tagger & 67.49  & 76.71  & 95.12  \\
    BERT-tagger-joint $\dag$ & 67.14  & 77.07  & 94.86  \\
    NNShot & 60.39  & 72.33  & 95.04  \\
    StructShot & 22.63  & 53.34  & 90.18  \\
    \midrule
    templateNER & 54.63  & 69.94  & 64.92  \\
    \midrule
    LST-NER w/o $\mathcal{D}_{gw}$+$\mathcal{L}_{aux}$ & 67.29  & 76.63  & 95.04  \\
    LST-NER w/o $\mathcal{L}_{aux}$ & 68.53  & 77.65  & 95.20  \\
    LST-NER w/o $\mathcal{D}_{gw}$ & 68.49  & 77.86  & 95.27  \\
    LST-NER (Ours) & \textbf{70.25 } & \textbf{78.74 } & \textbf{95.41 } \\
    \bottomrule
    \end{tabular}%
    }
  \caption{Cross domain results on three different domains in rich-resource setting. $\dag$ indicates both source and target labeled samples are used when training.}
  \label{tab:rich_cross_domain_table}%
\end{table}%

\section{Experiments}
\subsection{Experimental Settings}

\textbf{Datasets.}\quad
We take five public publicly available datasets for experiments, including CoNLL 2003 \citep{DBLP:conf/conll/SangM03}, CrossNER  \citep{DBLP:conf/aaai/Liu0YDJCMF21}, ATIS  \citep{DBLP:conf/interspeech/Hakkani-TurTCCG16}, MIT Restaurant \citep{DBLP:conf/icassp/LiuPCG13} and MIT Movie \citep{DBLP:conf/asru/LiuPWCG13}.
Table \ref{tab:dataset_statistics} presents detailed statistics of these datasets.

\noindent\textbf{Baseline models.}\quad
We first consider three approaches built on bi-directional LSTM structure \citep{hochreiter1997long}, including traditional NER system BiLSTM-CRF \citep{DBLP:conf/naacl/LampleBSKD16} together with two improved methods Coach \citep{DBLP:conf/acl/LiuWXF20} and Multi-Cell LSTM \citep{DBLP:conf/acl/JiaZ20}. 

We also compare several BERT-based NER systems.
BERT-tagger \citep{DBLP:conf/naacl/DevlinCLT19} is the BERT-based baseline model which fine-tunes the BERT model with a label classifier (\textit{i.e.,} \textit{pretrain-finetune}). 
NNShot and StructShot \citep{DBLP:conf/emnlp/YangK20} are two metric-based few-shot learning approaches for NER. 
Different from the above approaches, TemplateNER \citep{DBLP:conf/acl/CuiWLYZ21} is a template-based prompt method through a generative pre-trained LM, BART \citep{DBLP:conf/acl/LewisLGGMLSZ20}, and it also shows effectiveness in few-shot NER.

In the experiments, we don't include approaches requiring extra unlabeled data for comparison, such as noisy supervised pre-training, self-training  \citep{huang2020few} and domain-adaptive pre-training \citep{DBLP:conf/aaai/Liu0YDJCMF21}.

\noindent\textbf{Implementation Details.}\quad
Throughout the experiments, we use BERT-based
model\cite{DBLP:conf/naacl/DevlinCLT19} as our backbone model.
The models were implemented in Pytorch \cite{paszke2019pytorch} on top of the BERT Huggingface implementation \cite{wolf2019huggingface}, and training was performed on two GeForce RTX 2080 Ti GPU.

The hyperparameters in our model are set as follows: temperature parameter $T=4$; dimensional parameters $d_h=d_p=768$; edge threshold $\delta=1.5$; weight parameters $\lambda_1=0.1,\lambda_2=0.01$.

\noindent\textbf{Evaluation.}\quad
For evaluation, we use the standard evaluation metrics for NER (\textit{i.e.,} micro averaged F1 score) and report the average results of five independent runs.
Besides, we use BIO tagging schema for evaluation.

In the low-resource setting, we construct the target domain training set by sampling $K$ entities for each entity types following existing studies in few-shot NER \citep{DBLP:conf/emnlp/YangK20,DBLP:conf/acl/CuiWLYZ21}.
Different from sentence-level few-shot tasks, in NER, simply sampling $K$ sentences for each entity type will result in far more entities of frequent types than those of less frequent types \citep{DBLP:conf/emnlp/YangK20}.
Therefore, we apply greedy sampling strategy \citep{DBLP:conf/emnlp/YangK20} to construct a few-shot training set.
Due to the randomness of few-shot sampling, we will release all sampled data along with the codes for reproducibility.

\subsection{Cross-Domain Experiments}
\textbf{Cross-Domain Settings.}\quad
Following \citet{huang2020few,DBLP:conf/aaai/Liu0YDJCMF21}, we use CoNLL 2003 as the source domain datasets and evaluate the cross-domain performance on other datasets with different domains.
The MIT Movie, MIT Restaurant, and ATIS are three NER benchmark datasets.
However, these three datasets lack domain-specialized entity types or do not focus on a specific domain (\textit{e.g.,} "Opinion", "Relationship",etc), leading to a less effective cross-domain evaluation \citep{DBLP:conf/aaai/Liu0YDJCMF21}.
Thus, we additionally use CrossNER datasets (with five different domains) for the experiments.
For each domain in CrossNER, it contains domain-specialized entity types as well as the four entity types in CoNLL 2003\footnote{person, location, organization and miscellaneous}.
Since the target domain contains far more entity types than the source domain, there is a mismatch between different domain label sets.
Considering the statistics of the datasets, we perform experiments on movie reviews, restaurant reviews, and dialogue domains for the rich-resource setting (we use all samples for training) and all eight domains for the low-resource setting ($K=20,50$).
If an entity has a smaller number of samples than the fixed number to sample $K$, we use all of them for training.

\noindent\textbf{Training Details.}\quad
Based on the two baseline methods BiLSTM-CRF and BERT-tagger, we jointly train on both source and target domain samples to obtain two more baselines (\textit{i.e.,} BiLSTM-CRF-joint and BERT-tagger-joint, respectively) for better comparison.
Following \citet{DBLP:conf/aaai/Liu0YDJCMF21}, we up-sample target domain samples for balancing two domain data. 
When training BiLSTM-CRF and Coach, we use word-level embedding from \citet{DBLP:conf/emnlp/PenningtonSM14} and char-level embedding from \citet{DBLP:conf/emnlp/HashimotoXTS17} as input.
For Multi-Cell LSTM, BERT representation, as well as word-level and char-level embedding, are utilized.

Apart from the approaches based on multi-task learning (\textit{i.e.,} BiLSTM-CRF-joint, Coach, Multi-Cell LSTM, and BERT-tagger-joint), we train the NER model on CoNLL 2003 for ten epochs before adapting to the target domain.
For NNShot and StructShot, we further perform fine-tuning in the target domain since we find that they only yield better results than fine-tuning when only very few data are available \citep{huang2020few}.
We summarize the results of cross-domain evaluation as well as the ablation study in Table \ref{tab:low_cross_domain_table} and \ref{tab:rich_cross_domain_table}, where methods are grouped together based on the backbone model (BiLSTM, BERT, BART from top to down respectively).

\noindent\textbf{Result Analysis.}\quad
Results show that our model consistently outperforms all the compared models in both low- and rich-resource settings.
Our method shows significant improvements in the rich-resource setting on the baseline BERT-tagger (2.76\% on Movie Review; 2.03\% on Restaurant Review; 0.29\% on Dialogue).
Even though the multi-task-learning-based methods (\textit{e.g.,} Multi-Cell LSTM) are trained on more data and show competitive results, the proposed method has superior performance with only target domain data.

Results also suggest that jointly training pre-trained LM (\textit{e.g.,} BERT) on both domains data may not have better performance on target domain compared with \textit{pretrain-finetune} paradigm. 
We think that the reason may be the semantic discrepancy of the same label from two domains.
Different from them, the proposed method captures both similarity and discrepancy between source and target labels through probability distributions.
Therefore, our model benefits from the broad knowledge from the source NER model and alleviates the requirement to target domain data.

\noindent\textbf{Ablation Study.}\quad
We consider three settings in the ablation study, the final loss without (1) loss of auxiliary task $\mathcal{L}_{aux}$, (2) GWD for graph matching $\mathcal{D}_{gw}$ and (3) both of them.
One should note that the model trained in case (3) is not the same as BERT-tagger, which has label semantic fusion layers.

The results suggest that both the graph matching mechanism and label semantic fusion are beneficial for learning a better NER model. 
When training only with classification loss, the model shows tiny improvement on fine-tuning.
Combined with learned graph structure (\textit{i.e.,} \textit{source graph}), the label semantic fusion part becomes more effective when auxiliary task is added.
Moreover, the model trained with graph matching consistently yields better results, indicating that transferring the graph structure of labels is critical and beneficial for cross-domain NER.

\begin{table}[htbp]
  \tiny
  \centering
    \begin{tabular}{lp{0.4cm}p{0.4cm}p{0.4cm}p{0.4cm}p{0.4cm}p{0.4cm}}
    \toprule
    \textbf{Domain} & \multicolumn{1}{l}{Poli.} & \multicolumn{1}{l}{Sci.} & \multicolumn{1}{l}{Mus.} & \multicolumn{1}{l}{Lit.} & AI    & \multicolumn{1}{l}{Aver.} \\
    \midrule
    BERT-tagger $\ddag$ & 68.71  & 64.94  & 68.30  & 63.63  & 58.88  & 64.89  \\
    DAPT $\ddag$ & 72.05  & 68.78  & 75.71  & 69.04  & 62.56  & 69.63  \\
    Multi-Cell LSTM $\ddag$ & 70.56  & 66.42  & 70.52  & 66.96  & 58.28  & 66.55  \\
    Multi-Cell LSTM+DAPT $\ddag$ & 71.45  & 67.68  & 74.19  & 68.63  & 61.64  & 68.72  \\
    LST-NER (Ours) & 70.44  & 66.83  & 72.08  & 67.12  & 60.32  & 67.36  \\
    LST-NER+DAPT & \textbf{73.25 } & \textbf{70.07 } & \textbf{76.83 } & \textbf{70.76 } & \textbf{63.28 } & \textbf{70.84 } \\[-7pt]
    \bottomrule
    \end{tabular}%
  \caption{Comparison of different methods combined with DAPT. In each domain, we use \textbf{all} samples for training. $\ddag$ indicates the results are from \citet{DBLP:conf/aaai/Liu0YDJCMF21}.}
  \label{tab:dapt_table}%
\end{table}%

\subsection{Additional Experiments}
\noindent\textbf{Combined with Domain-Adaptive Pre-Training.}\quad
\citet{DBLP:conf/aaai/Liu0YDJCMF21} proposed to use integrate the entity-
and task-level unlabeled corpus and span-level masking strategy in Domain-Adaptive Pre-Training (DAPT) for the NER domain adaptation.
We conduct experiments to combine DAPT with ours model and Multi-Cell LSTM, respectively.
The results are shown in Table \ref{tab:dapt_table}.

By pre-training on a massive domain-related corpus, our method further improves the F1-score by 3.48\% on average.
Compared with Multi-Cell LSTM, our method benefits from rich knowledge learned by pre-train LM directly and shows better performance when combined with DAPT.
Therefore, we believe that our method can be incorporated with self-training and noisy supervised pre-training methods to achieve superior results. 

\begin{figure}[htbp]
    \centering
    \includegraphics[width=\linewidth]{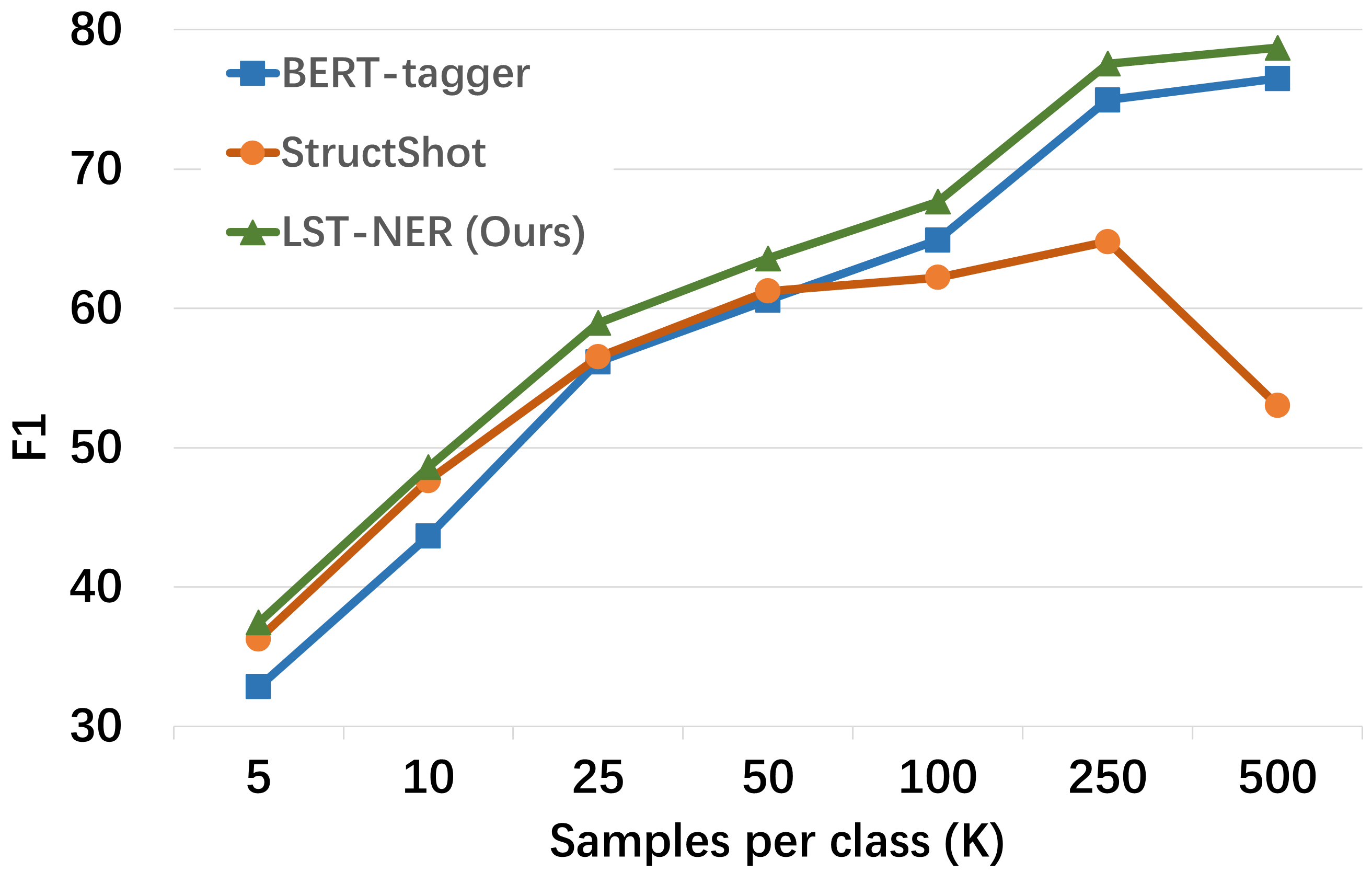}
    \caption{Comparisons when utilizing different amounts of data for training in "Restaurant Reviews" domain.}
    \label{fig:fig3}
\end{figure}

\noindent\textbf{Performance with Different Amounts of Data.}\quad
We evaluate the performance of our model with different amounts of target domain labeled data on the "Restaurant Reviews" domain and make comparisons with two baselines BERT-tagger and StructShot.
We use the same few-shot sampling strategy as in the low-resource setting.
From results in Fig \ref{fig:fig3}, we find that even when in a highly low-resource scenario ($K=5,10$), the proposed model shows competitive performance with the few-shot NER model StructShot.
When more data are available, our model consistently outperforms both BERT-tagger and StructShot.
In contrast, StructShot becomes ineffective when data are relatively sufficient (K>50).
We think the reason may be that StructShot is based on nearest neighbor learning, which is susceptible to noisy data. 
The results indicate that our method enhances domain adaptation capability in a more general scenario compared with few-shot NER methods.

\begin{figure}[htbp]
    \centering
    \includegraphics[width=\linewidth]{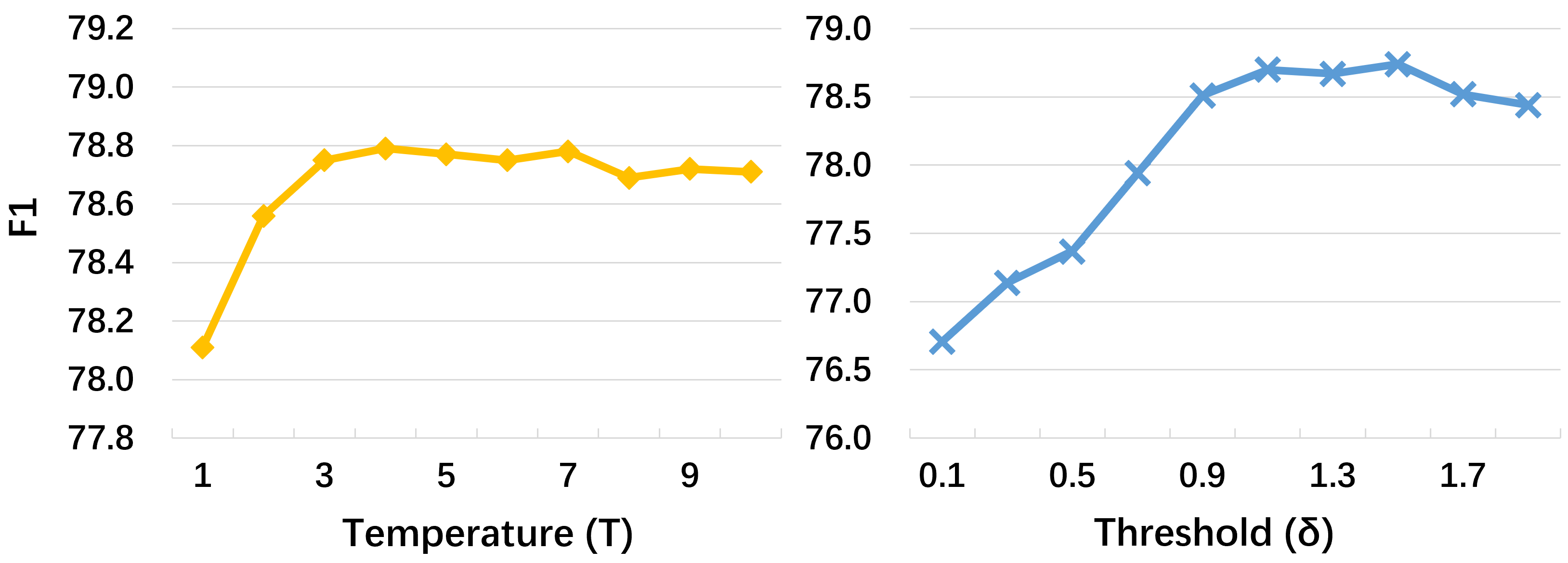}
    \caption{The impact of temperature $T$ and edge threshold $\delta$ to the performance in "Restaurant Reviews" domain.}
    \label{fig:fig4}
\end{figure}
\begin{figure}[htbp]
    \centering
    \includegraphics[width=\linewidth]{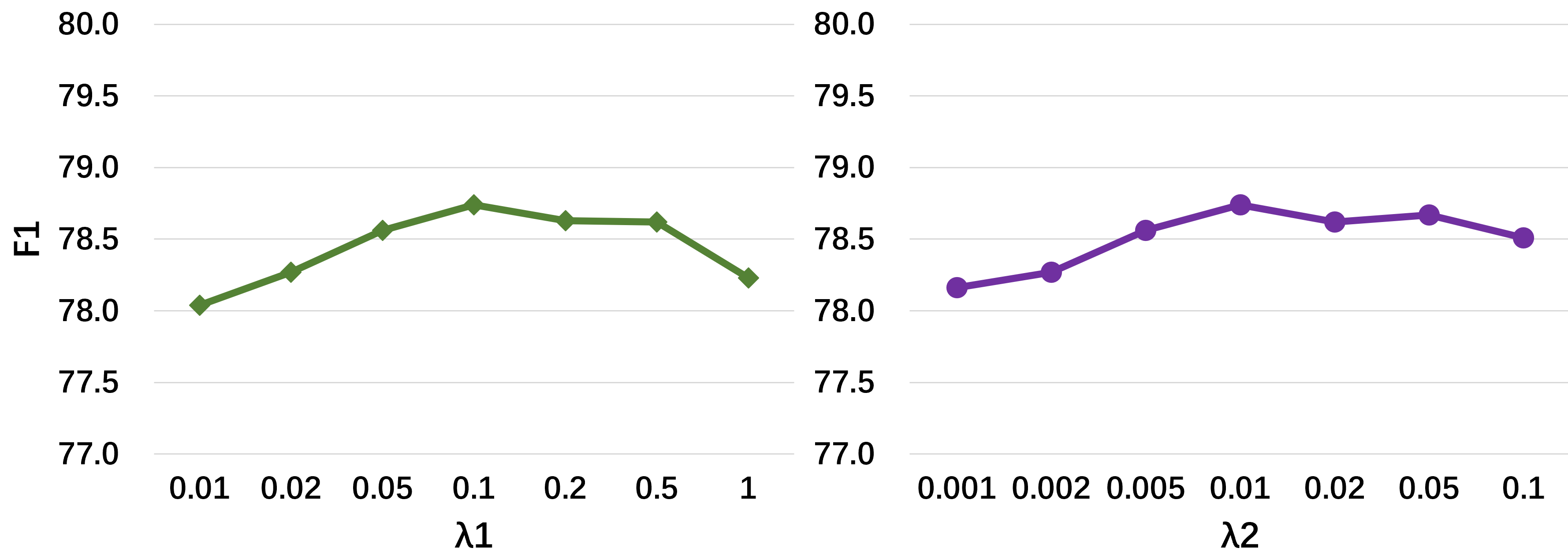}
    \caption{The impact of weight parameters $\lambda_1$ and $\lambda_2$ to the performance in "Restaurant Reviews" domain.}
    \label{fig:fig7}
\end{figure}

\noindent\textbf{Hyperparameter Discussion.}\quad
We explore the impact of edge threshold $\delta$, temperature parameter $T$ and weight parameter $\lambda_1$,$\lambda_2$ on the performance.
We show the result in Fig. \ref{fig:fig4} and Fig. \ref{fig:fig7}.
Temperature $T$ controls the smoothness of the score distribution. 
The edge threshold $\delta$ controls the number of edges for matching.
We find that $T$ and $\delta$ have a relatively small influence on the f1 score when $T>3$ and $\delta>1.0$, suggesting the stability of our model. 
In the experiments, we choose the best value as the default setting (\textit{i.e.,} $T=4$, $\delta=1.5$, $\lambda_1=0.1$ and $\lambda_2=0.01$).

\begin{figure}[h]
    \centering
    \subfloat[Example]{
    \includegraphics[width=\linewidth]{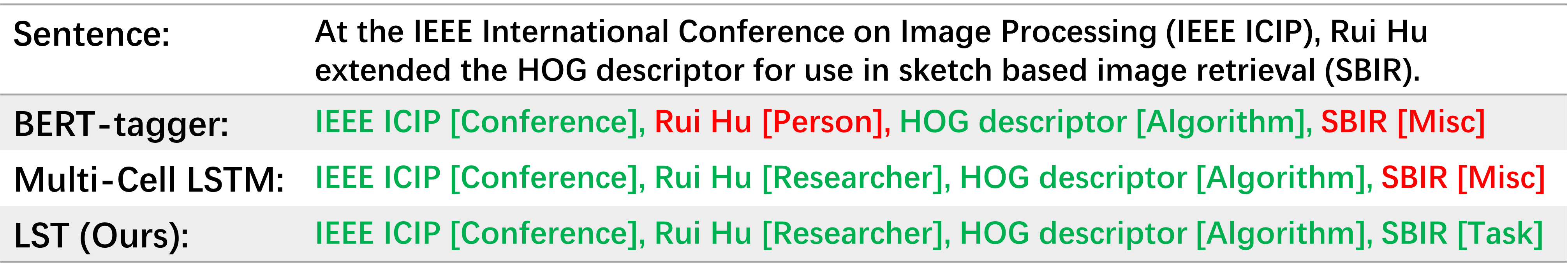}
    }
    \quad
    \subfloat[Transport plan]{
    \includegraphics[width=\linewidth]{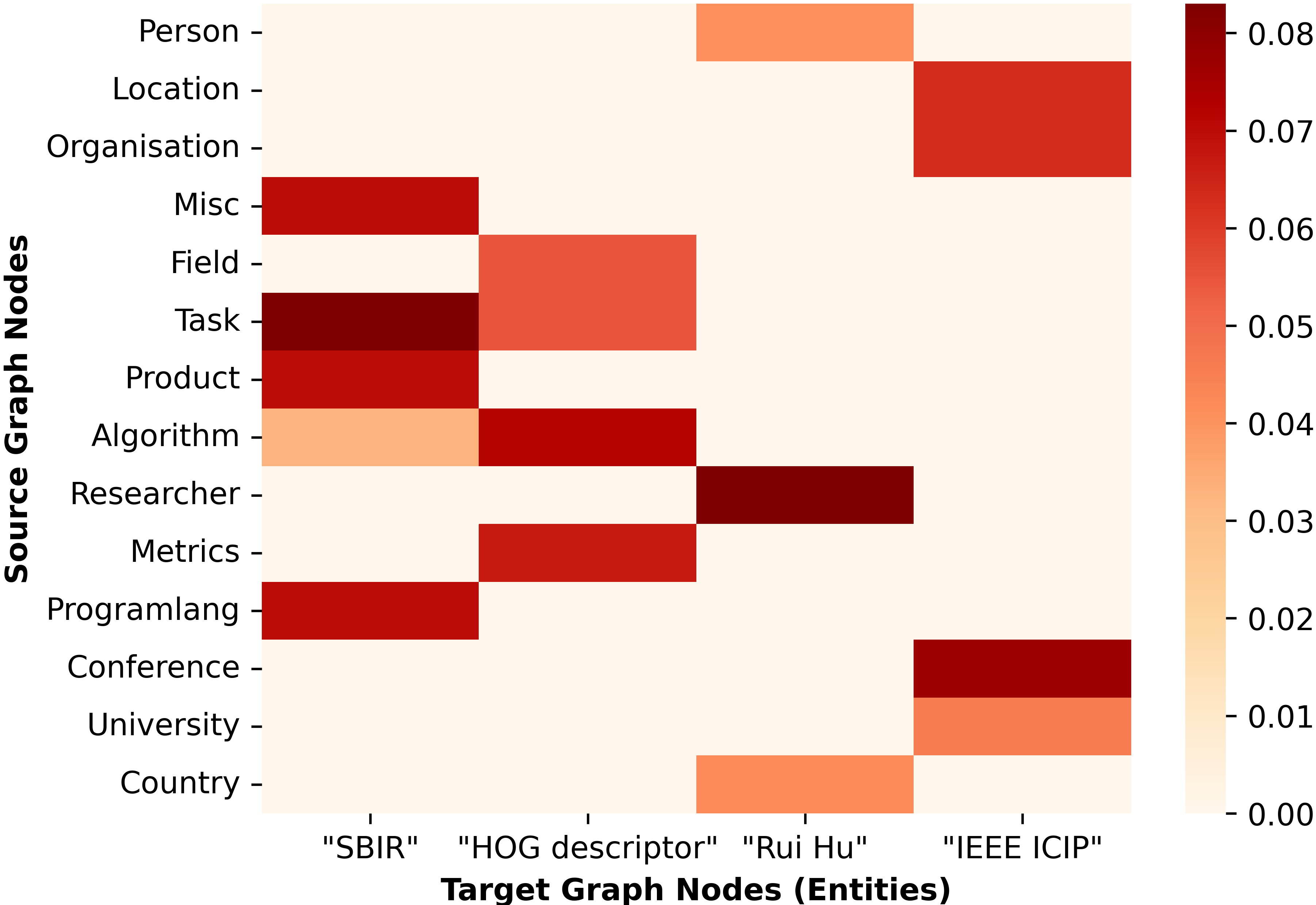}
    }
    \caption{(a) An example from the AI domain test set. Green and Red represent correct and incorrect entity respectively. (b) The transport plan corresponds to the example. A higher value represents more attention between nodes.}
    \label{fig:fig5}
\end{figure}

\noindent\textbf{Case Study.}\quad
In the example shown in Fig.\ref{fig:fig5}, the model constructs \textit{source graph} with all target data where all target labels are contained.
The transport plan demonstrates how label structures (edges) are learned via graph matching from all target entity types to the named entity in the sentence.
Compared with BERT-tagger and Multi-Cell LSTM, our method correctly predicts "Rui Hu" as "Researcher" and "SBIR" as "Task".

\section{Conclusion}
This paper proposes a novel and lightweight transfer learning approach for cross-domain NER.
Our proposed method learns graph structure via matching \textit{label graphs} from source to target domain.
Through extensive experiments, we demonstrated the effectiveness of our approach, reporting better results over a series of transfer learning, multi-task learning, few-shot learning methods.
In conclusion, our approach is general, which can be combined with domain-adaptive pre-training and potentially applied to other cross-domain prediction tasks.
Besides, there are some limitations of our approach. 
For example, when the target domain entity types are fine-grained and largely different from the source domain entity types (e.g., in ATIS dataset), our approach shows limited improvement on the \textit{pretrain-finetune} paradigm.
To this end, future directions include investigations on employing multi-task learning for modeling the semantic discrepancy of labels across domains and fusing hierarchical label relationships into the \textit{label graphs}.

\section*{Acknowledgments}
The work described in this paper was partially funded by the National Natural Science Foundation of China (Grant Nos. 61502174, 61872148), the Natural Science Foundation of Guangdong Province (Grant Nos. 2017A030313355, 2019A1515010768, 2021A1515011496), the Guangzhou Science and Technology Planning Project (Grant No. 201902010020), and the Fundamental Research Funds for the Central Universities.

\bibliography{anthology,custom}

\begin{thebibliography}{41}
\expandafter\ifx\csname natexlab\endcsname\relax\def\natexlab#1{#1}\fi

\bibitem[{Alvarez{-}Melis and
  Jaakkola(2018)}]{DBLP:conf/emnlp/Alvarez-MelisJ18}
David Alvarez{-}Melis and Tommi~S. Jaakkola. 2018.
\newblock Gromov-wasserstein alignment of word embedding spaces.
\newblock In \emph{Proceedings of the Conference on Empirical Methods in
  Natural Language Processing}, pages 1881--1890.

\bibitem[{Chen et~al.(2020)Chen, Gan, Cheng, Li, Carin, and
  Liu}]{DBLP:conf/icml/ChenG0LC020}
Liqun Chen, Zhe Gan, Yu~Cheng, Linjie Li, Lawrence Carin, and Jingjing Liu.
  2020.
\newblock Graph optimal transport for cross-domain alignment.
\newblock In \emph{Proceedings of the International Conference on Machine
  Learning}.

\bibitem[{Chowdhury and M{\'e}moli(2019)}]{chowdhury2019gromov}
Samir Chowdhury and Facundo M{\'e}moli. 2019.
\newblock The gromov--wasserstein distance between networks and stable network
  invariants.
\newblock \emph{Information and Inference: A Journal of the IMA},
  8(4):757--787.

\bibitem[{Cui et~al.(2021)Cui, Wu, Liu, Yang, and
  Zhang}]{DBLP:conf/acl/CuiWLYZ21}
Leyang Cui, Yu~Wu, Jian Liu, Sen Yang, and Yue Zhang. 2021.
\newblock Template-based named entity recognition using {BART}.
\newblock In \emph{Findings of the Association for Computational Linguistics},
  volume {ACL/IJCNLP} 2021 of \emph{Findings of {ACL}}, pages 1835--1845.

\bibitem[{Cuturi(2013)}]{DBLP:conf/nips/Cuturi13}
Marco Cuturi. 2013.
\newblock Sinkhorn distances: Lightspeed computation of optimal transport.
\newblock In \emph{Advances in Neural Information Processing Systems}, pages
  2292--2300.

\bibitem[{Devlin et~al.(2019)Devlin, Chang, Lee, and
  Toutanova}]{DBLP:conf/naacl/DevlinCLT19}
Jacob Devlin, Ming{-}Wei Chang, Kenton Lee, and Kristina Toutanova. 2019.
\newblock {BERT:} pre-training of deep bidirectional transformers for language
  understanding.
\newblock In \emph{Proceedings of the Conference of the North American Chapter
  of the Association for Computational Linguistics: Human Language
  Technologies}.

\bibitem[{Feng et~al.(2018)Feng, Feng, Qin, Feng, and
  Liu}]{DBLP:conf/ijcai/FengF0F018}
Xiaocheng Feng, Xiachong Feng, Bing Qin, Zhangyin Feng, and Ting Liu. 2018.
\newblock Improving low resource named entity recognition using cross-lingual
  knowledge transfer.
\newblock In \emph{Proceedings of the International Joint Conference on
  Artificial Intelligence}, pages 4071--4077.

\bibitem[{Hakkani{-}T{\"{u}}r et~al.(2016)Hakkani{-}T{\"{u}}r, T{\"{u}}r,
  Celikyilmaz, Chen, Gao, Deng, and
  Wang}]{DBLP:conf/interspeech/Hakkani-TurTCCG16}
Dilek Hakkani{-}T{\"{u}}r, G{\"{o}}khan T{\"{u}}r, Asli Celikyilmaz, Yun{-}Nung
  Chen, Jianfeng Gao, Li~Deng, and Ye{-}Yi Wang. 2016.
\newblock Multi-domain joint semantic frame parsing using bi-directional
  {RNN-LSTM}.
\newblock In \emph{Interspeech Annual Conference of the International Speech
  Communication Association}, pages 715--719.

\bibitem[{Hashimoto et~al.(2017)Hashimoto, Xiong, Tsuruoka, and
  Socher}]{DBLP:conf/emnlp/HashimotoXTS17}
Kazuma Hashimoto, Caiming Xiong, Yoshimasa Tsuruoka, and Richard Socher. 2017.
\newblock A joint many-task model: Growing a neural network for multiple {NLP}
  tasks.
\newblock In \emph{Proceedings of the 2017 Conference on Empirical Methods in
  Natural Language Processing7}, pages 1923--1933.

\bibitem[{Hochreiter and Schmidhuber(1997)}]{hochreiter1997long}
Sepp Hochreiter and J{\"u}rgen Schmidhuber. 1997.
\newblock Long short-term memory.
\newblock \emph{Neural computation}, 9(8):1735--1780.

\bibitem[{Houlsby et~al.(2019)Houlsby, Giurgiu, Jastrzebski, Morrone,
  de~Laroussilhe, Gesmundo, Attariyan, and
  Gelly}]{DBLP:conf/icml/HoulsbyGJMLGAG19}
Neil Houlsby, Andrei Giurgiu, Stanislaw Jastrzebski, Bruna Morrone, Quentin
  de~Laroussilhe, Andrea Gesmundo, Mona Attariyan, and Sylvain Gelly. 2019.
\newblock Parameter-efficient transfer learning for {NLP}.
\newblock In \emph{Proceedings of the International Conference on Machine
  Learning}, Proceedings of Machine Learning Research.

\bibitem[{Huang et~al.(2020)Huang, Li, Subudhi, Jose, Balakrishnan, Chen, Peng,
  Gao, and Han}]{huang2020few}
Jiaxin Huang, Chunyuan Li, Krishan Subudhi, Damien Jose, Shobana Balakrishnan,
  Weizhu Chen, Baolin Peng, Jianfeng Gao, and Jiawei Han. 2020.
\newblock Few-shot named entity recognition: A comprehensive study.
\newblock \emph{arXiv preprint arXiv:2012.14978}.

\bibitem[{Jia et~al.(2019)Jia, Xiao, and Zhang}]{DBLP:conf/acl/JiaXZ19}
Chen Jia, Liang Xiao, and Yue Zhang. 2019.
\newblock Cross-domain {NER} using cross-domain language modeling.
\newblock In \emph{Proceedings of the Conference of the Association for
  Computational Linguistics}, pages 2464--2474.

\bibitem[{Jia and Zhang(2020)}]{DBLP:conf/acl/JiaZ20}
Chen Jia and Yue Zhang. 2020.
\newblock Multi-cell compositional {LSTM} for {NER} domain adaptation.
\newblock In \emph{Proceedings of the Annual Meeting of the Association for
  Computational Linguistics}, pages 5906--5917.

\bibitem[{Kipf and Welling(2017)}]{DBLP:conf/iclr/KipfW17}
Thomas~N. Kipf and Max Welling. 2017.
\newblock Semi-supervised classification with graph convolutional networks.
\newblock In \emph{5th International Conference on Learning Representations,
  {ICLR} 2017, Toulon, France, April 24-26, 2017, Conference Track
  Proceedings}.

\bibitem[{Lample et~al.(2016)Lample, Ballesteros, Subramanian, Kawakami, and
  Dyer}]{DBLP:conf/naacl/LampleBSKD16}
Guillaume Lample, Miguel Ballesteros, Sandeep Subramanian, Kazuya Kawakami, and
  Chris Dyer. 2016.
\newblock Neural architectures for named entity recognition.
\newblock In \emph{The Conference of the North American Chapter of the
  Association for Computational Linguistics: Human Language Technologies},
  pages 260--270.

\bibitem[{Lee et~al.(2018)Lee, Dernoncourt, and
  Szolovits}]{DBLP:conf/lrec/LeeDS18}
Ji~Young Lee, Franck Dernoncourt, and Peter Szolovits. 2018.
\newblock Transfer learning for named-entity recognition with neural networks.
\newblock In \emph{Proceedings of the International Conference on Language
  Resources and Evaluation}.

\bibitem[{Lewis et~al.(2020)Lewis, Liu, Goyal, Ghazvininejad, Mohamed, Levy,
  Stoyanov, and Zettlemoyer}]{DBLP:conf/acl/LewisLGGMLSZ20}
Mike Lewis, Yinhan Liu, Naman Goyal, Marjan Ghazvininejad, Abdelrahman Mohamed,
  Omer Levy, Veselin Stoyanov, and Luke Zettlemoyer. 2020.
\newblock {BART:} denoising sequence-to-sequence pre-training for natural
  language generation, translation, and comprehension.
\newblock In \emph{Proceedings of the Annual Meeting of the Association for
  Computational Linguistics}, pages 7871--7880.

\bibitem[{Lin and Lu(2018)}]{DBLP:conf/emnlp/LinL18}
Bill~Yuchen Lin and Wei Lu. 2018.
\newblock Neural adaptation layers for cross-domain named entity recognition.
\newblock In \emph{Proceedings of the Conference on Empirical Methods in
  Natural Language Processing}, pages 2012--2022.

\bibitem[{Liu et~al.(2013{\natexlab{a}})Liu, Pasupat, Cyphers, and
  Glass}]{DBLP:conf/icassp/LiuPCG13}
Jingjing Liu, Panupong Pasupat, Scott Cyphers, and James~R. Glass.
  2013{\natexlab{a}}.
\newblock Asgard: {A} portable architecture for multilingual dialogue systems.
\newblock In \emph{{IEEE} International Conference on Acoustics, Speech and
  Signal Processing}, pages 8386--8390.

\bibitem[{Liu et~al.(2013{\natexlab{b}})Liu, Pasupat, Wang, Cyphers, and
  Glass}]{DBLP:conf/asru/LiuPWCG13}
Jingjing Liu, Panupong Pasupat, Yining Wang, Scott Cyphers, and James~R. Glass.
  2013{\natexlab{b}}.
\newblock Query understanding enhanced by hierarchical parsing structures.
\newblock In \emph{{IEEE} Workshop on Automatic Speech Recognition and
  Understanding}, pages 72--77.

\bibitem[{Liu et~al.(2020{\natexlab{a}})Liu, Winata, and
  Fung}]{DBLP:conf/rep4nlp/LiuWF20}
Zihan Liu, Genta~Indra Winata, and Pascale Fung. 2020{\natexlab{a}}.
\newblock Zero-resource cross-domain named entity recognition.
\newblock In \emph{Proceedings of the Workshop on Representation Learning for
  NLP}, pages 1--6.

\bibitem[{Liu et~al.(2020{\natexlab{b}})Liu, Winata, Xu, and
  Fung}]{DBLP:conf/acl/LiuWXF20}
Zihan Liu, Genta~Indra Winata, Peng Xu, and Pascale Fung. 2020{\natexlab{b}}.
\newblock Coach: {A} coarse-to-fine approach for cross-domain slot filling.
\newblock In \emph{Proceedings of the Annual Meeting of the Association for
  Computational Linguistics}, pages 19--25.

\bibitem[{Liu et~al.(2021)Liu, Xu, Yu, Dai, Ji, Cahyawijaya, Madotto, and
  Fung}]{DBLP:conf/aaai/Liu0YDJCMF21}
Zihan Liu, Yan Xu, Tiezheng Yu, Wenliang Dai, Ziwei Ji, Samuel Cahyawijaya,
  Andrea Madotto, and Pascale Fung. 2021.
\newblock Crossner: Evaluating cross-domain named entity recognition.
\newblock In \emph{Thirty-Fifth {AAAI} Conference on Artificial Intelligence,
  {AAAI} 2021, Thirty-Third Conference on Innovative Applications of Artificial
  Intelligence, {IAAI} 2021, The Eleventh Symposium on Educational Advances in
  Artificial Intelligence, {EAAI} 2021, Virtual Event, February 2-9, 2021},
  pages 13452--13460.

\bibitem[{Nadeau and Sekine(2007)}]{nadeau2007survey}
David Nadeau and Satoshi Sekine. 2007.
\newblock A survey of named entity recognition and classification.
\newblock \emph{Lingvisticae Investigationes}, 30(1):3--26.

\bibitem[{Paszke et~al.(2019)Paszke, Gross, Massa, Lerer, Bradbury, Chanan,
  Killeen, Lin, Gimelshein, Antiga et~al.}]{paszke2019pytorch}
Adam Paszke, Sam Gross, Francisco Massa, Adam Lerer, James Bradbury, Gregory
  Chanan, Trevor Killeen, Zeming Lin, Natalia Gimelshein, Luca Antiga, et~al.
  2019.
\newblock Pytorch: An imperative style, high-performance deep learning library.
\newblock \emph{Advances in neural information processing systems}, 32.

\bibitem[{Pennington et~al.(2014)Pennington, Socher, and
  Manning}]{DBLP:conf/emnlp/PenningtonSM14}
Jeffrey Pennington, Richard Socher, and Christopher~D. Manning. 2014.
\newblock Glove: Global vectors for word representation.
\newblock In \emph{Proceedings of the Conference on Empirical Methods in
  Natural Language Processing}, pages 1532--1543.

\bibitem[{Peyr{\'{e}} et~al.(2016)Peyr{\'{e}}, Cuturi, and
  Solomon}]{DBLP:conf/icml/PeyreCS16}
Gabriel Peyr{\'{e}}, Marco Cuturi, and Justin Solomon. 2016.
\newblock Gromov-wasserstein averaging of kernel and distance matrices.
\newblock In \emph{Proceedings of the 33nd International Conference on Machine
  Learning}, volume~48, pages 2664--2672.

\bibitem[{Peyr{\'e} et~al.(2019)Peyr{\'e}, Cuturi
  et~al.}]{peyre2019computational}
Gabriel Peyr{\'e}, Marco Cuturi, et~al. 2019.
\newblock Computational optimal transport: With applications to data science.
\newblock \emph{Foundations and Trends{\textregistered} in Machine Learning},
  11(5-6):355--607.

\bibitem[{Rodr{\'{\i}}guez et~al.(2018)Rodr{\'{\i}}guez, Caldwell, and
  Liu}]{DBLP:conf/coling/RodriguezCL18}
Juan~Diego Rodr{\'{\i}}guez, Adam Caldwell, and Alexander Liu. 2018.
\newblock Transfer learning for entity recognition of novel classes.
\newblock In \emph{Proceedings of the International Conference on Computational
  Linguistics}, pages 1974--1985.

\bibitem[{Sang and Meulder(2003)}]{DBLP:conf/conll/SangM03}
Erik F. Tjong~Kim Sang and Fien~De Meulder. 2003.
\newblock Introduction to the conll-2003 shared task: Language-independent
  named entity recognition.
\newblock In \emph{Proceedings of the Conference on Natural Language Learning},
  pages 142--147.

\bibitem[{Shaalan(2014)}]{DBLP:journals/coling/Shaalan14}
Khaled Shaalan. 2014.
\newblock A survey of arabic named entity recognition and classification.
\newblock \emph{Computational Linguistics}, 40(2):469--510.

\bibitem[{Tong et~al.(2021)Tong, Wang, Xu, Cao, Liu, Hou, and
  Li}]{tong-etal-2021-learning}
Meihan Tong, Shuai Wang, Bin Xu, Yixin Cao, Minghui Liu, Lei Hou, and Juanzi
  Li. 2021.
\newblock Learning from miscellaneous other-class words for few-shot named
  entity recognition.
\newblock In \emph{Proceedings of the Annual Meeting of the Association for
  Computational Linguistics and the International Joint Conference on Natural
  Language Processing}, pages 6236--6247.

\bibitem[{van Lint and Wilson(1992)}]{DBLP:books/daglib/0073486}
Jacobus~H. van Lint and Richard~M. Wilson. 1992.
\newblock \emph{A course in combinatorics}.
\newblock Cambridge University Press.

\bibitem[{Vayer et~al.(2019)Vayer, Courty, Tavenard, Chapel, and
  Flamary}]{DBLP:conf/icml/VayerCTCF19}
Titouan Vayer, Nicolas Courty, Romain Tavenard, Laetitia Chapel, and R{\'{e}}mi
  Flamary. 2019.
\newblock Optimal transport for structured data with application on graphs.
\newblock In \emph{Proceedings of the 36th International Conference on Machine
  Learning}, volume~97, pages 6275--6284.

\bibitem[{Wang et~al.(2020)Wang, Kulkarni, and
  Preotiuc{-}Pietro}]{DBLP:conf/acl/WangKP20}
Jing Wang, Mayank Kulkarni, and Daniel Preotiuc{-}Pietro. 2020.
\newblock Multi-domain named entity recognition with genre-aware and agnostic
  inference.
\newblock In \emph{Proceedings of the Annual Meeting of the Association for
  Computational Linguistics}, pages 8476--8488.

\bibitem[{Wolf et~al.(2019)Wolf, Debut, Sanh, Chaumond, Delangue, Moi, Cistac,
  Rault, Louf, Funtowicz et~al.}]{wolf2019huggingface}
Thomas Wolf, Lysandre Debut, Victor Sanh, Julien Chaumond, Clement Delangue,
  Anthony Moi, Pierric Cistac, Tim Rault, R{\'e}mi Louf, Morgan Funtowicz,
  et~al. 2019.
\newblock Huggingface's transformers: State-of-the-art natural language
  processing.
\newblock \emph{arXiv preprint arXiv:1910.03771}.

\bibitem[{Yang and Katiyar(2020)}]{DBLP:conf/emnlp/YangK20}
Yi~Yang and Arzoo Katiyar. 2020.
\newblock Simple and effective few-shot named entity recognition with
  structured nearest neighbor learning.
\newblock In \emph{Proceedings of the Conference on Empirical Methods in
  Natural Language Processing}, pages 6365--6375.

\bibitem[{Yang et~al.(2017)Yang, Salakhutdinov, and
  Cohen}]{DBLP:conf/iclr/YangSC17}
Zhilin Yang, Ruslan Salakhutdinov, and William~W. Cohen. 2017.
\newblock Transfer learning for sequence tagging with hierarchical recurrent
  networks.
\newblock In \emph{International Conference on Learning Representations}.

\bibitem[{You et~al.(2020)You, Kou, Long, and Wang}]{DBLP:conf/nips/YouKL020}
Kaichao You, Zhi Kou, Mingsheng Long, and Jianmin Wang. 2020.
\newblock Co-tuning for transfer learning.
\newblock In \emph{Advances in Neural Information Processing Systems 33: Annual
  Conference on Neural Information Processing Systems}.

\bibitem[{Zheng et~al.(2022)Zheng, Liang, Chen, and Ma}]{zheng2022distilling}
Junhao Zheng, Zhanxian Liang, Haibin Chen, and Qianli Ma. 2022.
\newblock Distilling causal effect from miscellaneous other-class for continual
  named entity recognition.
\newblock In \emph{Proceedings of the 2022 Conference on Empirical Methods in
  Natural Language Processing}, pages 3602--3615.

\end{thebibliography}
\bibliographystyle{acl_natbib}

\end{document}